\documentclass[conference]{IEEEtran}

\usepackage{graphics} 
\graphicspath{ {figures/} }
\usepackage{caption}
\usepackage{subcaption}
\usepackage{epsfig} 
\usepackage{amsmath} 
\usepackage[ruled]{algorithm2e} 
\usepackage{url}
\usepackage{bm} 
\usepackage{comment}
\usepackage{amssymb}
\usepackage{multirow}
\usepackage{tabularx}
\usepackage{cite}
\usepackage{color,soul}
\newcolumntype{C}[1]{>{\centering\arraybackslash}p{#1}}

\begin{document}
%
\title{Sim2Real2Sim: Bridging the Gap Between Simulation and Real-World in Flexible Object Manipulation}

\author{\IEEEauthorblockN{Peng Chang}
\IEEEauthorblockA{College of Engineering\\
Northeastern University\\
Boston, Massachusetts, USA\\
Email: chang.pe@husky.neu.edu}
\and
\IEEEauthorblockN{Ta\c{s}k{\i}n~Pad{\i}r}
\IEEEauthorblockA{Institute for Experiential Robotics\\
Northeastern University\\
Boston, Massachusetts, USA\\
Email: t.padir@northeastern.edu}}
\maketitle
\begin{abstract}
This paper addresses a new strategy called Simulation-to-Real-to-Simulation (Sim2Real2Sim) to bridge the gap between simulation and real-world, and automate a flexible object manipulation task. This strategy consists of three steps: (1) using the rough environment with the estimated models to develop the methods to complete the manipulation task in the simulation; (2) applying the methods from simulation to real-world and comparing their performance; (3) updating the models and methods in simulation based on the differences between the real world and the simulation. The Plug Task from the 2015 DARPA Robotics Challenge Finals is chosen to evaluate our Sim2Real2Sim strategy. A new identification approach for building the model of the linear flexible objects is derived from real-world to simulation. The automation of the DRC plug task in both simulation and real-world proves the success of the Sim2Real2Sim strategy. Numerical experiments are implemented to validate the simulated model.
\end{abstract}

\IEEEpeerreviewmaketitle

\section{Introduction}


Simulation plays a very important role in robotics \cite{vzlajpah2008simulation}. Various simulators (e.g., Gazebo, OpenRAVE, MuJoCo) are used to design the robot model, create different simulation environment, analyze the kinematics and dynamics of the robot, design different plan or control algorithms, investigate the performances of the system, etc. The construction of a real environment is usually more expensive than building a simulated environment. In simulation, we can build objects with their geometric and physical properties. We can also design new robot models or import existing robot models to complete a series of manipulation tasks. Usually the robot needs to make physical contact with the environment and objects when completing the manipulation tasks. Learning to use the robot to manipulate objects in simulation can help with avoiding damage to the robot, the environment, and the humans. 
\begin{figure}[ht]
\vspace{2mm}
    \centering
    \includegraphics[width=\linewidth]{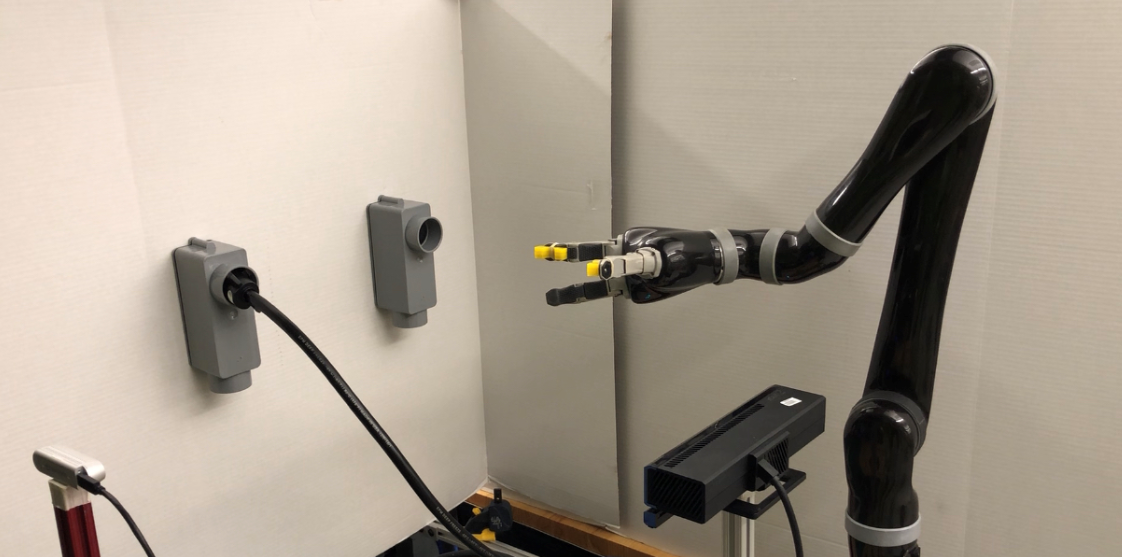}
    \caption{The DRC Plug Task setup in real-world.}
    \label{fig:real}
    \vspace{-2mm}
\end{figure}

Manipulation of flexible objects is generally more difficult than manipulating rigid objects because flexible objects have higher order and more complex dynamics. To predict the deformation of a flexible object, it is required to model it with accurate physical properties. Mass-spring \cite{essahbi2012soft} and Finite-element \cite{muller2002stable} methods are two popular methods for modeling flexible objects. Yoshida et al. \cite{yoshida2015simulation} modeled an $O$-ring with the finite-element method and assembled this ring-shape object on a rigid cylinder in simulation. Alvarez and Yamazaki \cite{alvarez2016interactive} modeled a linear flexible object with a physics engine and proposed an interactive way to change the simulation environment. However, their work have only been implemented in simulation as the object models are greatly simplified \cite{sanchez2018robotic}.
There are also approaches not relying on the physical model of the flexible objects. Navarro-Alarcon and Liu \cite{navarro2018fourier} proposed a feedback method to represent the object's shape based on a truncated Fourier series, and the robotic arm could deform the flexible objects to desired shapes in the 2-D plane with this visual model. Zhu et al. \cite{zhu2018dual} applied the approach in \cite{navarro2018fourier} and realized the shape control of a flexible cable in the 2-D plane. In these works, the physical models of the objects are not necessary because the objects are confined to two grippers and the deformation of the objects are only in the 2-D plane. In our case, The DRC Plug Task involves manipulating a cable that deforms due to gravity. So, a physical model is required for this task to estimate the deformation, and we plan to use the simulation to help with modeling the cable. Figure~\ref{fig:real} shows the plug task setup in real-world. The details of the setup will be discussed in the next section.

The gap between simulation and real-world was introduced in Simulation-to-Real (Sim2Real) \cite{farchy2013humanoid, lund1996simulated, koos2010crossing, carpin2006bridging}. Simulation is observed to have inevitable simplifications with heavy optimization. Besides, there exist physical events not modeled in simulators and parameters of the simulated models that need to be identified. Therefore, policies or parameters learned \cite{hanna2017bridging} and controllers designed \cite{koos2010crossing} in simulation can not be transferred to the real world directly.

\begin{figure}[ht]
\vspace{2mm}
    \centering
    \includegraphics[width=.9\linewidth]{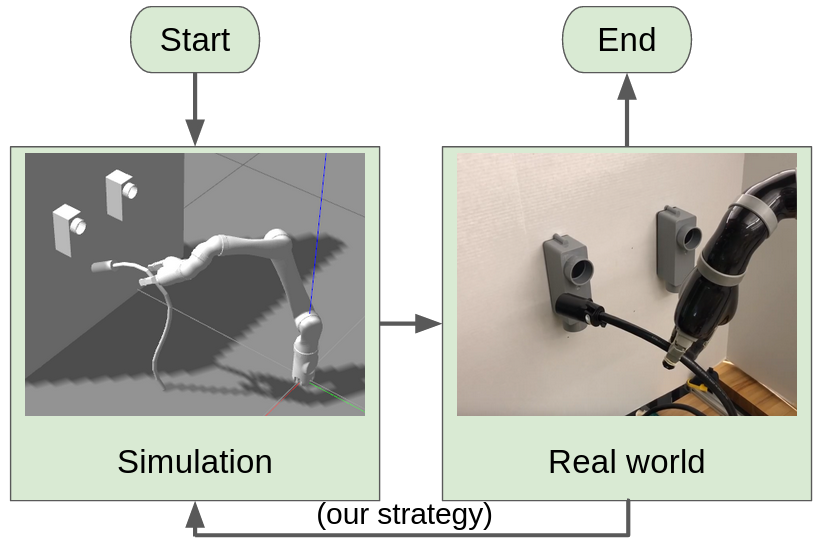}
    \caption{Sim2Real2Sim flowchart representation.}
    \label{fig:Sim2Real2Sim}
\vspace{-2mm}
\end{figure}

To reduce the gap between simulation and real-world, we update the simulation environment based on the real-world data. Figure~\ref{fig:Sim2Real2Sim} shows the flowchart of the Sim2Real2Sim strategy we will use for bridging the gap. We start with a rough simulation environment with the estimated models. Then we test the system framework in the real world based on the methods developed in simulation and collect the data from the real world. Finally, we go back to the simulation and update the models and methods based on the data from the real world. This paper makes the following contributions: (1) it introduces a novel strategy Sim2Real2Sim for bridging the gap between simulation and real-world; (2) it presents a system identification approach of getting the physical model of linear flexible objects. The overall result is a high-fidelity simulation environment for flexible object manipulation. 

This paper is organized as follows. Section II presents the problem statement. Section III describes the methodology of the Sim2Real2Sim strategy applied to the DRC Plug Task and the cable model identification approach. In Section IV, we show the results of using Sim2Real2Sim strategy and compare the model of the cable in simulation with the real cable. Section V includes the conclusion and discussion of possible future work.

\section{PROBLEM STATEMENT}
The DARPA Robotics Challenge (DRC) is a competition funded by the US Defense Advanced Research Projects Agency between 2012 and 2015. The DARPA Robotics Challenge (DRC) Simulator (DRCSim) and the Space Robotics Challenge Simulation (SRCSim) have related open-source simulations for most tasks in the competition. However, the Plug Task used in the 2015 DARPA Robotics Challenge Finals does not have the complete simulation environment. The DRC Plug Task requires the robot to pull the cable (with a cylindrical plug) out of one socket and plug it into another socket \cite{spenko2018darpa}. Based on our group's prior experience in this competition \cite{dedonato2017team} and the detailed definition and requirements of this task, we decided to build a complete simulation environment and automate this task for our research study. 

A task setup is shown in Fig.~\ref{fig:real}, based on the requirements listed in \cite{spenko2018darpa}. The setup consists of a wall, two power sockets, and a power cable with a plug. A Kinova Jaco V2 robotic arm and two RGB-D cameras are used to complete this task. One challenge in this task is the modeling of the cable, including geometric and physical models. Another challenge is how to narrow the gap between the simulated model and the real cable. Moreover, automating this task also presents a challenge. In order to automate the DRC Plug Task, we separate this task into five phases: \textsc{Initialize}, \textsc{Grasp}, \textsc{Unplug}, \textsc{Pre-insert}, and \textsc{Insert}. In the \textsc{Initialize} phase, the robot should sense the environment. After getting the position of the cable, the robot should move to \textsc{Grasp} the cable, then \textsc{Unplug} the cable. We specify a \textsc{Pre-insert} phase to make the robot move the cable-tip in front of the target socket and perform the \textsc{Insert} phase better.

\section{METHODOLOGY}
Research approach - from simulation to real-world implementation is widely applied in humanoid robot \cite{wonsick2019analysis}, mobile robot \cite{peng2018sim}, robotic arm \cite{bousmalis2018using}, etc. Normally, we complete a task in simulation first, then transfer the methods used in simulation to the real world. This step is named as ``simulation to real-world'' in the Sim2Real2Sim strategy. Building a simulation environment with all rigid objects is well-developed and these objects have nearly the same behavior in simulation and real-world. But simulation of soft objects usually simplifies the model, the gap between simulation and real-world can not be ignored. Our proposed Sim2Real2Sim approach is for solving this issue. More specifically, we use the real-world data to improve the simulated models and we call this step as ``real-world to simulation''.

\subsection{Simulation to real-world}



\begin{figure}[ht]
\vspace{-1mm}
    \centering
    \includegraphics[width=.5\linewidth]{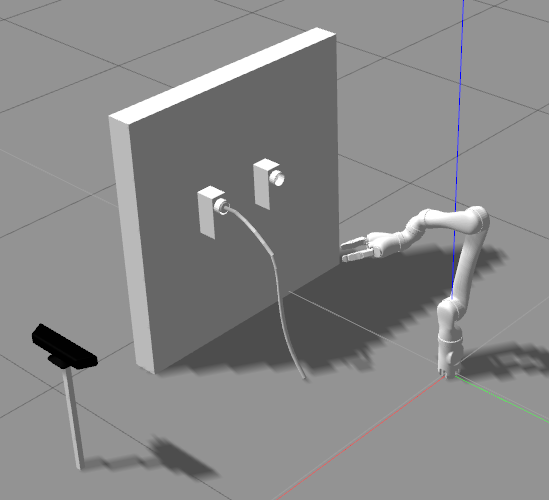}
    \caption{Simulation environment.}
    \label{fig:gazebo}
\vspace{-1mm}
\end{figure}

Led by the Sim2Real2Sim strategy, we build an initial simulation environment based on the task requirements. Figure~\ref{fig:gazebo} shows the simulation setup in the Gazebo simulator \cite{koenig2004design}. Modeling of rigid objects in the simulator is a solved problem: the wall can be modeled as a box and the plug can be fitted as a cylinder. Modeling objects of irregular shape is relatively difficult, but we can get the model in a solid modeling computer-aided design tool (e.g., SolidWorks) and then use a converter (e.g., SolidWorks to URDF Exporter) to get the model in the desired format. The RGB-D camera in the simulated environment is a Kinect camera. To read the point cloud information, we need to model the Kinect camera as a Depth Camera Plugin in Gazebo. The simulated Jaco arm is cloned from the official Kinova GitHub website. In simulation, we assume that the positions of the sockets are known.

\begin{figure}[ht]%
\centering
\begin{subfigure}[b]{0.45\linewidth}
\includegraphics[width=\textwidth, height=.8\textwidth]{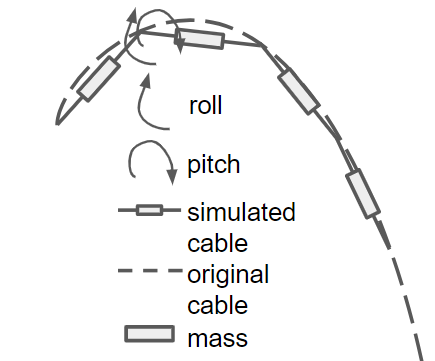}%
\caption[a]{Augmented representation.}
\label{fig:cable_model-a}%
\end{subfigure}
\begin{subfigure}[b]{0.45\linewidth}
\includegraphics[width=\textwidth, height=.8\textwidth]{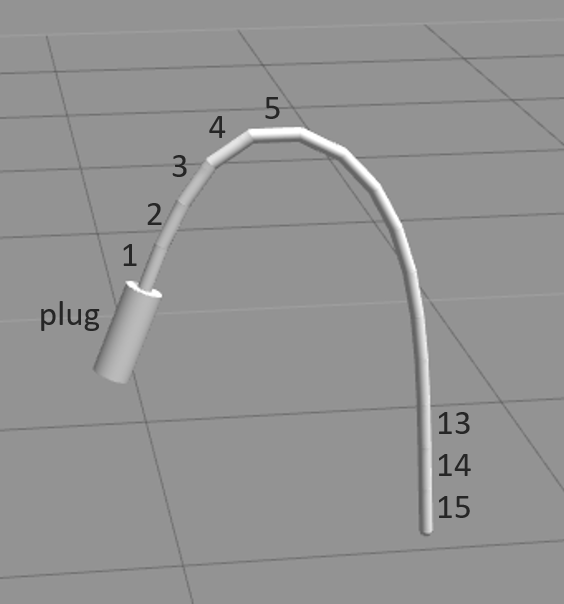} 
\caption[b]{Cable model in Gazebo.}
\label{fig:cable_model-b}%
\end{subfigure}
\caption{Cable model design.}%
\label{fig:cable_model}%
\end{figure}

Inspired by the ``fire hose" model in Gazebo and the Piecewise Constant Curvature (PCC) model \cite{webster2010design}, a Dynamically-Consistent Augmented Formulation \cite{della2018dynamic} of the PCC cable model is created in Fig.~\ref{fig:cable_model-a}. A Gazebo model of the cable is also built based on the PCC model (see Fig.~\ref{fig:cable_model-b}). Fifteen links are used to model the cable, each link is 5~$cm$ long and weighs 50~$g$ with the center of mass in the middle of the arc. The plug is also a link with length 10~$cm$ and weight 100~$g$. These links are connected by revolute joints (except the joint between the plug task and link-1) with the roll and pitch rotations. Each joint has stiffness and damping properties. 

\begin{figure}[ht]%
\vspace{2mm}
\centering
\begin{subfigure}[b]{0.45\linewidth}
\includegraphics[width=\textwidth, height=.8\textwidth]{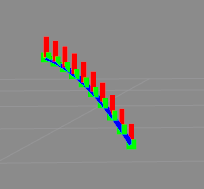}
\end{subfigure}
\begin{subfigure}[b]{0.45\linewidth}
\includegraphics[width=\textwidth, height=.8\textwidth]{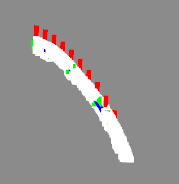}
\end{subfigure}
\caption{Geometric model of the cable. Green squares represent the sampling points, adjacent points are connected by blue lines, and vertical red lines are marked to illustrate the locations of the sampling points.}%
\label{fig:modeling_gazebo}%
\vspace{-4mm}
\end{figure}

Recall that the DRC Plug Task can be divided into five phases. First, we use the simulator to complete the \textsc{Grasp} operation. In simulation, different random points are given as the target poses to MoveIt (with the RRTConnect planner) \cite{chitta2012moveit}. 30 successful experiments prove that we can apply the same motion planning method to the real robot, and as expected, the real robot successfully grasps the cable with measured given points on the cable. After we use these given points for testing, we must figure out a way to detect the position of the cable from the camera. Modeling of the cable becomes our first choice. We find that two projections from the 3D curve to the 2D plane can fit the curve of this cable. A quadratic polynomial equation is used to fit the 2D curve in each plane and the least-squares method is used to get the coefficients of these two polynomial equations. By sampling along the common axis in these two 2D planes, we can get the geometric model of the cable. For example, if the two equations are $x=f(y)$ and $z=f(y)$, we can get the 3D points on the cable by sampling along the $y$ axis. This geometric model works both in simulation and real-world. A working example is shown in Fig.~\ref{fig:modeling_gazebo}. With the geometric model of the cable, the robot can \textsc{Grasp} the cable autonomously.

\begin{figure}[ht]%
\centering
\begin{subfigure}[b]{0.45\linewidth}
\includegraphics[width=\textwidth, height=.8\textwidth]{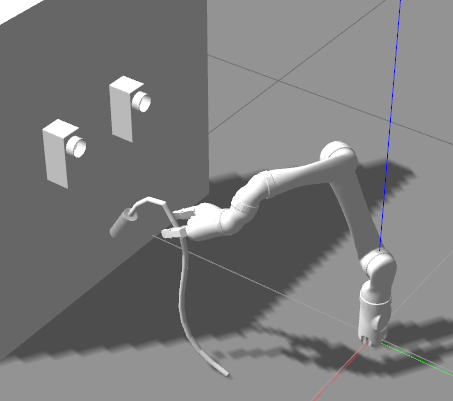}%
\caption[a]{Cable dangles too much.}
\label{fig:dangle-a}%
\end{subfigure}
\begin{subfigure}[b]{0.45\linewidth}
\includegraphics[width=\textwidth, height=.8\textwidth]{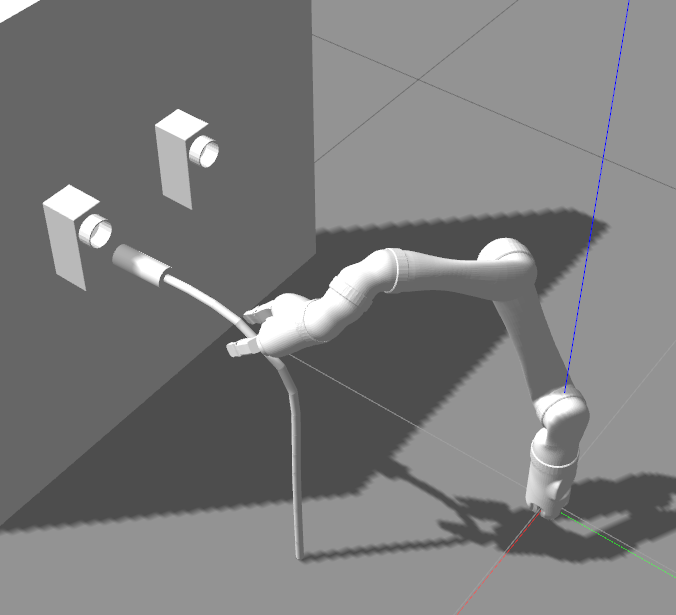} 
\caption[b]{Cable barely dangles.}
\label{fig:dangle-b}%
\end{subfigure}
\caption{Cable deformation in Gazebo.}%
\label{fig:cable_dangle}%
\end{figure}

\begin{figure}[ht]%
\centering
\begin{subfigure}[b]{0.45\linewidth}
\includegraphics[width=\textwidth, height=.8\textwidth]{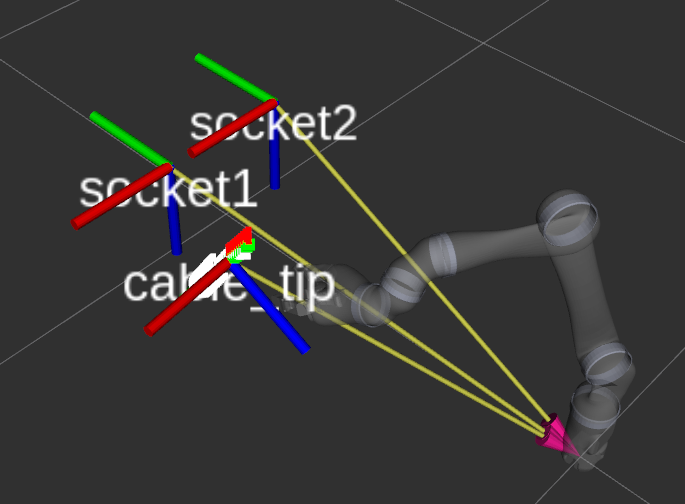}%
\caption[a]{Geometric model of the cable when it dangles too much.}
\label{fig:dangle-model-a}%
\end{subfigure}
\begin{subfigure}[b]{0.45\linewidth}
\includegraphics[width=\textwidth, height=.8\textwidth]{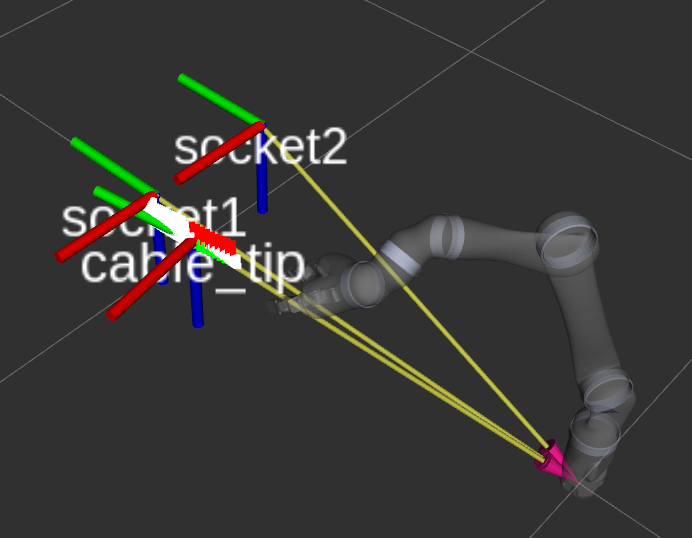} 
\caption[b]{Geometric model of the cable when it barely dangles.}
\label{fig:dangle-model-b}%
\end{subfigure}
\caption{Geometric model in Rviz.}%
\label{fig:cable_dangle_rviz}%
\end{figure}

\textsc{Unplugging} the cable from the socket is straightforward with a backward panning motion for the end-effector. After the cable being plugged out of the socket, the weight of the front section of the cable will cause the cable to dangle. In our model (see Fig.~\ref{fig:cable_model}), this deformation depends on the joint stiffness and damping. We observe that the deformation of the cable is different by randomly setting the values for the joint stiffness and damping. Two special cases are: (1) the cable dangles too much (see Fig.~\ref{fig:dangle-a}); (2) the cable barely dangles (see Fig.~\ref{fig:dangle-b}). However, our geometric model works well in both cases (see Fig.~\ref{fig:cable_dangle_rviz}). To get reasonable values for the joint stiffness and damping and to narrow the gap between simulation and real-world, we need to collect data from the real-world and go back to simulation to update the models and methods.

\vspace{-3mm}

\subsection{Real-world to simulation}


After implementing the methods developed in simulation to real-world, we realized that the geometric model is capable of estimating the pose of the cable. However, the deformation of the cable in simulation and real-world is different after the robot unplugs the cable from the socket, which means our physical model in simulation needs to be tuned to match with the real world.

As we mentioned before, the cable is modeled as a rigid manipulator with passive revolute joints (see Fig.~\ref{fig:cable_model}). The dynamics of the cable can be represented as:
\begin{equation} \label{eq:inverse_dynamics}
M\mathbf{\ddot{q}}+C\mathbf{\dot{q}}+G+J^T\mathbf{f}_{ext}+K\mathbf{q}+D\mathbf{\dot{q}}=\bm{\tau},
\end{equation}

where $\mathbf{q},\mathbf{\dot{q}},\mathbf{\ddot{q}}$ represent joint position, velocity, and acceleration. $M$ is the inertia matrix. $C$ is the centrifugal and Coriolis forces matrix. $G$ is gravitational forces or torques. $J^T$ is the transpose of the robot Jacobian. $\mathbf{f}_{ext}$ is the external force. $K$ is the stiffness and $D$ is the damping. $\bm{\tau}$ is a vector of joint torques, corresponding to the torques and forces applied by the actuators at the joints. Our goal is to estimate the stiffness and damping in this equation.
\begin{figure}[ht]%
\centering
\begin{subfigure}[b]{0.45\linewidth}
\includegraphics[width=\textwidth, height=.8\textwidth]{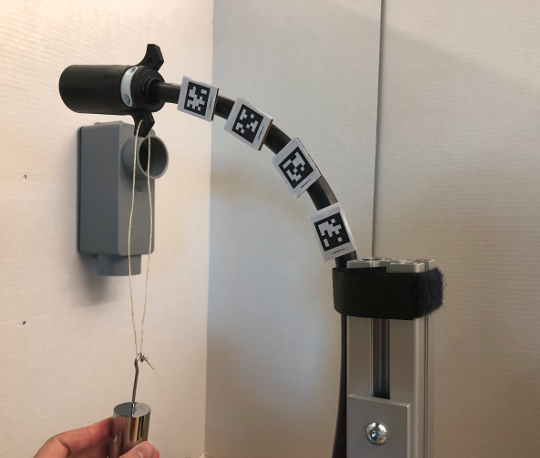}%
\caption[a]{Using apriltag to record the joint positions - start point.}
\label{fig:apriltag-a}%
\end{subfigure}
\begin{subfigure}[b]{0.45\linewidth}
\includegraphics[width=\textwidth, height=.8\textwidth]{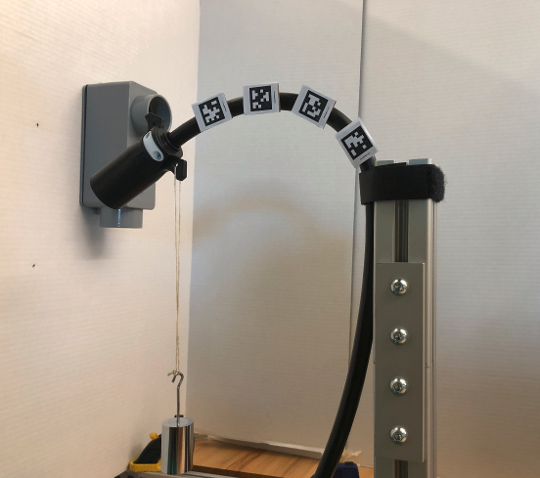} 
\caption[b]{Using apriltag to record the joint positions - end point.}
\label{fig:apriltag-b}%
\end{subfigure}
\caption{Apriltags attached on the cable for recording the joint values.}%
\label{fig:apriltag}%
\vspace{-2mm}
\end{figure}

Recalling that each joint of the cable has roll and pitch rotations (see Fig.~\ref{fig:cable_model}), we simplify the model by only considering the pitch. Our first challenge is to get the joint values. AprilTag \cite{olson2011apriltag} is used for getting the joint positions. Figure~\ref{fig:apriltag} shows the start and end points of recording the joint positions with a 100~$g$ weight added to the tip. Four apriltags with side lengths of 2.0~$cm$ \cite{dos2015evaluation} are attached on the cable 5~$cm$ apart (same as the link length in Gazebo). The joint position $\mathbf{q}$ and the time stamp are recorded for calculating joint velocity $\mathbf{\dot{q}}$ and acceleration $\mathbf{\ddot{q}}$. After we get joint values, the robot Jacobian $J$ can be calculated. Because the joints on the cable are passive, $\bm{\tau}$ equals to $\mathbf{0}$. The external force $\mathbf{f}_{ext}$ is also known by multiplying the mass of the weight (0.1~$kg$) with the gravitational acceleration (9.8 $m/s^2$).

\begin{figure}[ht]
    \centering
    \includegraphics[width=.8\linewidth]{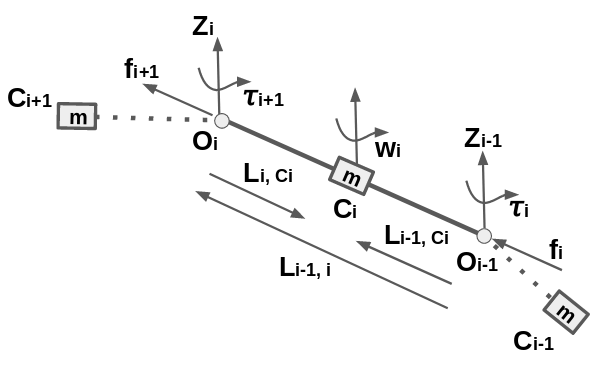}
    \caption{Parameters relating three adjacent links. Three links are $C_{i-1}$, $C_i$, and $C_{i+1}$. $w_i$ is the angular velocity of link $C_i$. $m$ is the mass of the link. $L_{i-1,i}$ is vector from the origin of frame $i-1$ to the origin of frame $i$. $L_{i-1,c_i}$ is vector from the origin of frame $i-1$ to the centre of mass $C_i$. $L_{i,c_i}$ is vector from the origin of frame $i$ to the centre of mass $C_i$. $z_i$ is vector points outward and is perpendicular to the paper. $f_i$ is force exerted by link $i-1$ on link $i$. $\tau_i$ is torque exerted by link $i-1$ to link $i$ with respect to origin of the frame $i-1$.} 
    \label{fig:rne}
\end{figure}

Computing of the left three terms ($M\mathbf{\ddot{q}}+C\mathbf{\dot{q}}+G$) in (\ref{eq:inverse_dynamics}) is our second challenge. We refer to the Recursive Newton Euler (RNE) approach \cite{walker1982efficient, luh1980line, siciliano2010robotics}. The RNE approach includes two recursions: forward recursion for computing velocities and accelerations, and backward recursion for computing forces and torques. Figure \ref{fig:rne} shows the parameters of three adjacent links and Algorithm \ref{algorithm:rne} shows the detailed RNE algorithm. The RNE algorithm gives us the torque required for each joint to overcome the gravity of the model and the force generated from the motion if the cable joints have no friction or damping and there is no external force. The left three terms ($M\mathbf{\ddot{q}}+C\mathbf{\dot{q}}+G$) in (\ref{eq:inverse_dynamics}) can be represented as: 
\begin{equation} \label{eq:tau}
M\mathbf{\ddot{q}}+C\mathbf{\dot{q}}+G=\bm{\tau}_{RNE},
\end{equation}
where $\bm{\tau}_{RNE}$ equals to $[\tau_1 \tau_2 ... \tau_n]^T$ ($n=4$) calculated from the RNE method.

\begin{algorithm}[]
 \textbf{Forward recursion:\{}\\
Initial: velocity and acceleration of the base link both equal to 0\\
Compute link angular velocity: $w_i=(R^{i-1}_i)^T (w_{i-1}+\dot{q}_i z_{i-1})$  \\
Compute link angular acceleration: $\dot{w}_i=(R^{i-1}_i)^T (\dot{w}_{i-1}+\ddot{q}_i z_{i-1}+\dot{q}_i w_{i-1}\times z_{i-1})$ \\
Compute linear acceleration of origin of frame $i$: $a_i=(R^{i-1}_i)^T a_{i-1} + \dot{w}_i\times L_{i-1,i}+w_i\times(w_i\times L_{i-1,i})$\\
Compute linear acceleration of centre of $C_i$: $a_{ci}=a_i+\dot{w}_i\times L_{i,c_i}+w_i\times(w_i\times L_{i,c_i})$
\}\\
 \textbf{Backward recursion:\{}\\
Compute force exerted by link $i-1$ on link $i$: $f_i=R^i_{i+1} f_{i+1}+m_i a_{ci}$\\
Compute moment exerted by link $i-1$ on link $i$: $\mu_i=-f_i\times (L_{i-1,i}+L_{i,c_i})+\mu_{i+1}+f_{i+1}\times L_{i,c_i}+I_i\dot{w}_i+w_i\times(I_i w_i)$\\
Compute torque exerted by link $i-1$ on link $i$: $\tau_i=\mu_i^T z_{i-1}+\tau_{inertia}$
\}\\
 \caption{Recursive Newton Euler algorithm}
 \label{algorithm:rne}
\end{algorithm}

\begin{figure}[ht]
    \centering
    \includegraphics[width=.6\linewidth]{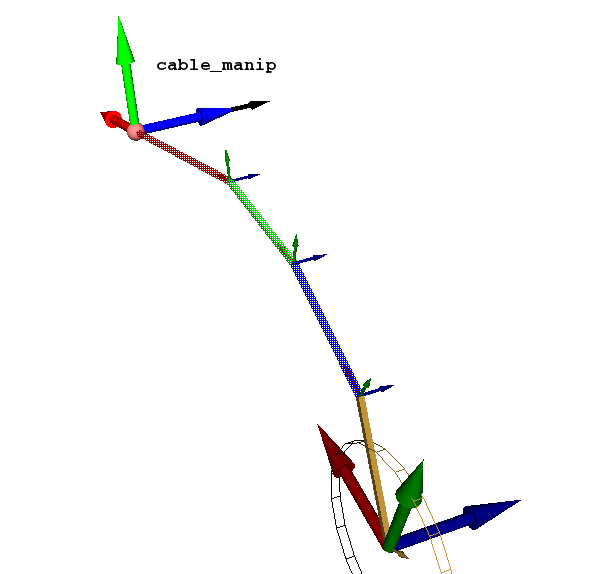}
    \caption{Cable model in OpenRAVE.} 
    \label{fig:openrave}
\end{figure}

OpenRAVE \cite{diankov2008openrave} is used for computing the robot Jacobian $J$ and the inverse dynamic terms $M\mathbf{\ddot{q}}+C\mathbf{\dot{q}}+G$. Figure~\ref{fig:openrave} shows the physical model built in OpenRAVE. We pick the first four links of the cable for modeling because the cable-tip pose is not guaranteed to be aligned with the socket pose if the grasp point is beyond this range. The robot Jacobian and inverse dynamics are computed based on this model.

At this point, we can rewrite (\ref{eq:inverse_dynamics}) as: 
\begin{equation} \label{eq:id_rewrite}
K\mathbf{q}+D\mathbf{\dot{q}}=\bm{\tau}-(M\mathbf{\ddot{q}}+C\mathbf{\dot{q}}+G+J^T\mathbf{f}_{ext}).
\end{equation}
The terms to the right of the equation (\ref{eq:id_rewrite}) are known. To get the stiffness $K$, we use the robot configuration when the joint velocity $\mathbf{\dot{q}}$ equals to $\mathbf{0}$. In our case, we use the joint values when the cable is finally stationary after adding the weight (see Fig.~\ref{fig:apriltag-b}). The stiffness $K$ can be computed by using (\ref{eq:stiffness}) with $\mathbf{\dot{q}}=\mathbf{0}$,
\begin{equation} \label{eq:stiffness}
K=(\bm{\tau}-(M\mathbf{\ddot{q}}+C\mathbf{\dot{q}}+G+J^T\mathbf{f}_{ext}+D\mathbf{\dot{q}}))\mathbf{q}^{+},
\end{equation}
where $\mathbf{q}^{+}$ is the pseudoinverse of $\mathbf{q}$.

After getting stiffness $K$, we take it to (\ref{eq:id_rewrite}). The damping $D$ can then be calculated as:
\begin{equation} \label{eq:damping}
D=(\bm{\tau}-(M\mathbf{\ddot{q}}+C\mathbf{\dot{q}}+G+J^T\mathbf{f}_{ext}+K\mathbf{q}))\mathbf{\dot{q}}^{+},
\end{equation}
where $\mathbf{\dot{q}}^{+}$ is the pseudo inverse of $\mathbf{\dot{q}}$. Unlike the calculation for stiffness, we use joint values prior to the cable entering its final stationary position to ensure that joint velocity is not $\mathbf{0}$.

\begin{figure}[ht]%
\vspace{2mm}
\centering
\begin{subfigure}[b]{0.45\linewidth}
\includegraphics[width=\textwidth, height=.8\textwidth]{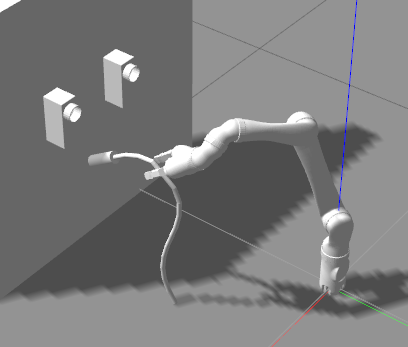}%
\caption[a]{Updated model in Gazebo.}
\label{fig:dangle_normal_gazebo}%
\end{subfigure}
\begin{subfigure}[b]{0.45\linewidth}
\includegraphics[width=\textwidth, height=.8\textwidth]{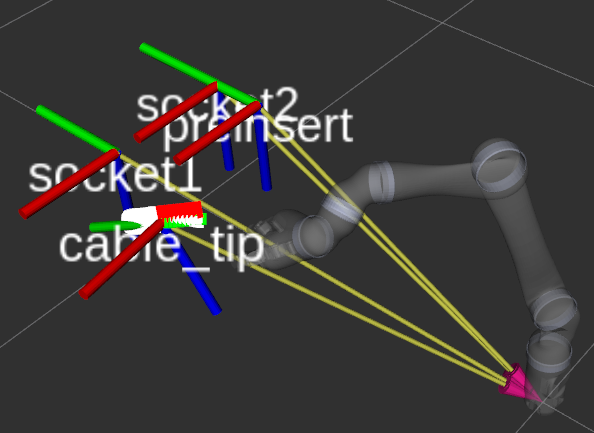} 
\caption[b]{Updated model in Rviz.}
\label{fig:dangle_normal_rviz}%
\end{subfigure}
\caption{Updated model with estimated joint stiffness and damping.}%
\label{fig:dangle_normal}%
\vspace{-4mm}
\end{figure}

Finally, we update the Gazebo model based on the calculated stiffness and damping. Recalling that each joint on the cable also has the roll rotation. We can use the same method to identify the stiffness and damping properties if we can figure out how to add a constant twist torque on the cable-tip. This part is left as a future work. Currently, we use the joint limits to set the roll rotation based on the observation. Figure~\ref{fig:dangle_normal} shows the deformation of the cable with the updated model parameters in the simulation. Detailed comparison of the model deformation in simulation with the cable deformation in real-world will be discussed in Section IV.

\subsection{Visual servoing in simulation and real-world}

To automate the Plug Task, we need to figure out how to carry the cable to the \textsc{Pre-insert} position (see Fig.~\ref{fig:dangle_normal_rviz}) after the cable being \textsc{Unplugged} from the original socket. Using motion planning to move the cable by taking the cable as one additional link of the robotic arm is not a good option due to the deformation of the cable. Because of the continuous visual feedback, we decide to use the visual servoing approach to align the cable-tip pose with the socket pose.

The visual servoing approach is also known as a vision-based robot control approach. The goal of this step is to reduce the differences between the ``cable\_tip'' frame and the \textsc{Pre-insert} frame (see Fig.~\ref{fig:dangle_normal_rviz}). Algorithm \ref{algorithm:visual_servoing} shows our visual servoing approach, which works both in simulation and real-world with different PID parameters. After the cable-tip moves to the \textsc{Pre-insert} position, a forward panning motion for the end-effector will \textsc{Insert} the cable-tip (plug) to the ``socket2'' (see Fig.~\ref{fig:dangle_normal_rviz}).

\begin{algorithm}[]
 \KwResult{``cable\_tip'' frame = \textsc{pre-insert} frame}
 \While{``cable\_tip'' frame != \textsc{pre-insert} frame}{
 Calculate the difference between the \textsc{Pre-insert} frame and the ``cable\_tip'' frame.\\
 Transfer the position difference to be Cartesian velocity ($\mathbf{v}$) by dividing by a constant time.\\
 Obtain the joint velocity ($\mathbf{\dot{q}}$) through Cartesian velocity ($\mathbf{v}$) and robot Jacobian ($J$): $\mathbf{\dot{q}}=J^{-1} \mathbf{v}$.\\
 Use PID controller to control the joint velocity.\\
 Send joint velocity to the robot.\\
 }
 \caption{Visual servoing approach}
 \label{algorithm:visual_servoing}
\end{algorithm}

\section{EXPERIMENTAL VALIDATION}
Our Sim2Real2Sim strategy successfully automates the DRC Plug Task, both in simulation and real-world. The performance is evaluated by running the task for 20 trials both in simulation and real-world. In simulation, the average completion time is 66.25~$s$ with an average real-time factor: 0.65. In real-world, the average completion time is 31.53~$s$. 
\begin{figure}[ht]%
\centering
\begin{subfigure}[b]{0.45\linewidth}
\includegraphics[width=\textwidth, height=.8\textwidth]{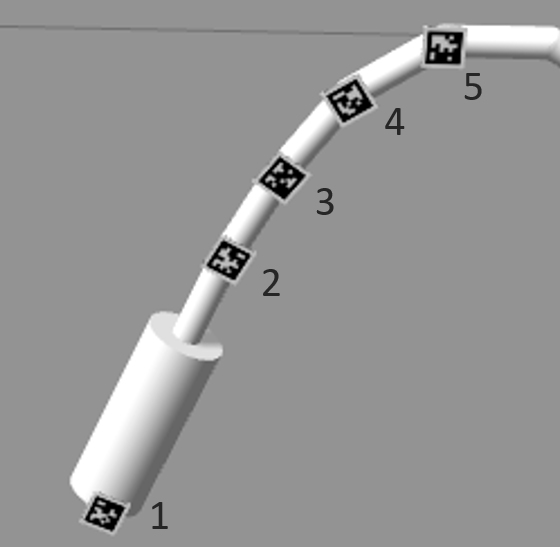}%
\caption[a]{Validation setup in simulation.}
\label{fig:validation_sim_gazebo}%
\end{subfigure}
\begin{subfigure}[b]{0.45\linewidth}
\includegraphics[width=\textwidth, height=.8\textwidth]{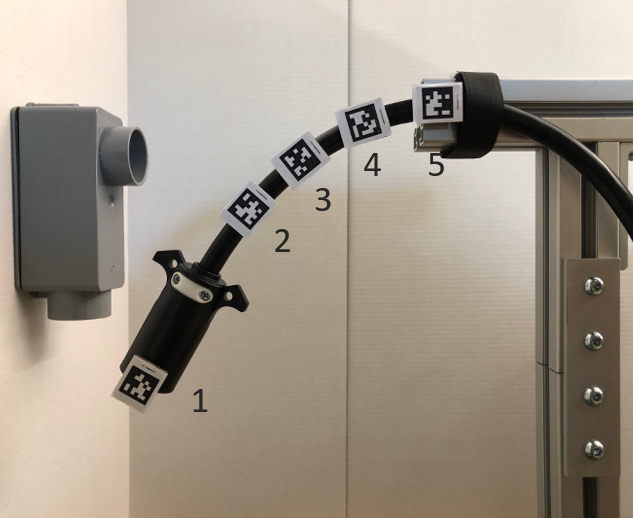} 
\caption[b]{Validation setup in real-world.}
\label{fig:validation_real_real}%
\end{subfigure}
\caption{Setup in simulation and real-world for comparison.}%
\label{fig:validation_setup}%
\vspace{-2mm}
\end{figure}

\begin{figure}[ht]%
\centering
\begin{subfigure}[b]{0.45\linewidth}
\includegraphics[width=\textwidth, height=.8\textwidth]{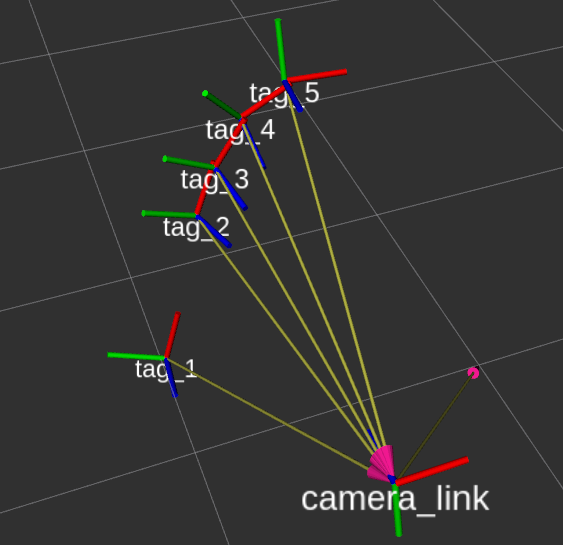}%
\caption[a]{Frames of the cable-tip and the revolute joints of the cable in simulation.}
\label{fig:validation_sim_rviz}%
\end{subfigure}
\begin{subfigure}[b]{0.45\linewidth}
\includegraphics[width=\textwidth, height=.8\textwidth]{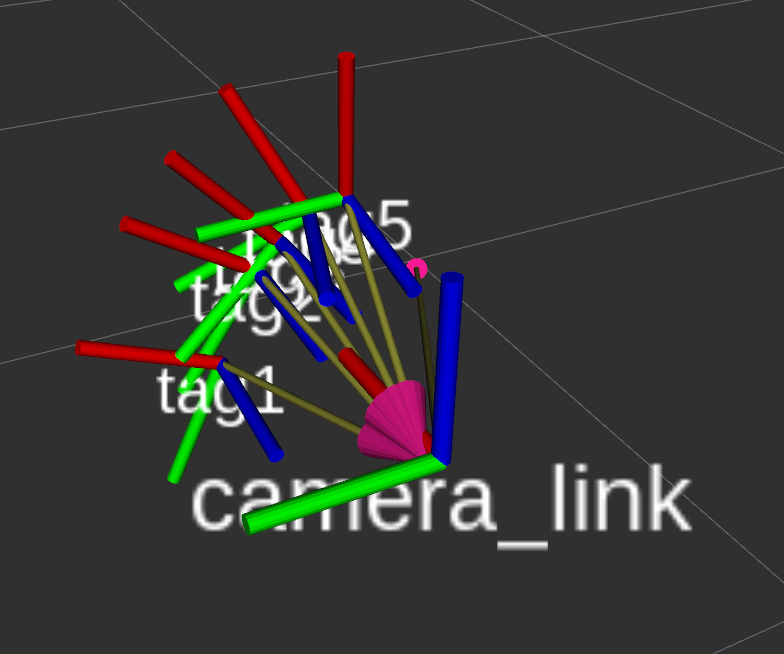} 
\caption[b]{Frames of the cable-tip and the revolute joints of the cable in real-world.}
\label{fig:validation_real_rviz}%
\end{subfigure}
\caption{Frames in simulation and real-world.}%
\label{fig:validation_rviz}%
\vspace{-2mm}
\end{figure}

To validate the model in simulation, we designed two experiments. First, we fix the same link on the cable to the world, and compare the joint positions of the cable in simulation and real-world. Second, we fix different links on the cable horizontally, and compare the pose of the cable-tip in simulation and real-world. In simulation, the link of the simulated cable can be fixed to the world by adding a ``fixed'' joint connecting the ``world'' link and the link you want to fix. In real-world, we fix the cable by using a zip-tie. Figure~\ref{fig:validation_setup} shows the setup in simulation and real-world for comparison of the cable deformation. Apriltags are attached to the cable tip and on the first four revolute joints to get the transformation at each joint. Figure~\ref{fig:validation_rviz} shows the visualized frames in Rviz, and the ``tag\_1'' frame is equivalent to the ``cable\_tip'' frame. 

Our first experiment is comparing the deformation of the cable by fixing the same link horizontally to the world. We use the setup in Fig.~\ref{fig:validation_setup} and obtain the frames (``tag\_1'', ``tag\_2'', ..., ``tag\_5'') both in simulation and real-world (see Fig.~\ref{fig:validation_rviz}). Transformations of these frames are used to get the joint positions for the comparison in simulation and real-world. We first compare the joint positions of the cable with no external force, then we add different weights to the cable-tip and compare the differences of the joint positions under these disturbances.

Table~\ref{table:joint_positions_no_mass} shows the comparison result by fixing the link-5 (the link after tag-5) on the cable to the world. From the last row of Table~\ref{table:joint_positions_no_mass}, we conclude that the shape of the cable in the simulation is close to the real cable with the largest difference of the joint positions being 0.005 $rad$, and the largest percent error between the simulation and the real-world is 2.00\%. To test the simulation model with external disturbances, we add two different weights (50~$g$ and 100~$g$) to the cable-tip, and we compare the joint positions of the simulated cable with the one of the real cable (see Table~\ref{table:joint_positions_50g} and Table~\ref{table:joint_positions_100g}). The percent errors are reasonable which proves the accuracy of our simulated cable model.

\begin{table}[ht]
\centering
\caption{Comparison of the joint positions in simulation and real-world (no external force)}
\label{table:joint_positions_no_mass}
\begin{tabular}{|c|c|c|c|c|}
\hline
Joint id / tag id              & 4      & 3      & 2      & 1      \\ \hline
Joint position in sim (rad)    & 0.560  & 0.387  & 0.168  & 0.098  \\ \hline
Joint position in real (rad)   & 0.565  & 0.383  & 0.171  & 0.100  \\ \hline
Difference (sim to real) (rad) & -0.005 & 0.004  & -0.003 & -0.002 \\ \hline
Percent error                  & 0.88\% & 1.04\% & 1.75\% & 2.00\% \\ \hline
\end{tabular}
\vspace{-2mm}
\end{table}

\begin{table}[ht]
\centering
\caption{Comparison of the joint positions in simulation and real-world (adding a 50~$g$ weight to the tip)}
\label{table:joint_positions_50g}
\begin{tabular}{|c|c|c|c|c|}
\hline
Joint id / tag id              & 4      & 3      & 2      & 1      \\ \hline
Joint position in sim (rad)    & 0.569  & 0.438  & 0.164  & 0.115  \\ \hline
Joint position in real (rad)   & 0.565  & 0.437  & 0.170  & 0.117  \\ \hline
Difference (sim to real) (rad) & 0.004  & 0.001  & -0.006 & -0.002 \\ \hline
Percent error                  & 0.71\% & 0.23\% & 3.53\% & 1.71\% \\ \hline
\end{tabular}
\vspace{-2mm}
\end{table}

\begin{table}[ht]
\centering
\caption{Comparison of the joint positions in simulation and real-world (adding a 100~$g$ weight to the tip)}
\label{table:joint_positions_100g}
\begin{tabular}{|c|c|c|c|c|}
\hline
Joint id / tag id              & 4      & 3      & 2      & 1      \\ \hline
Joint position in sim (rad)    & 0.619  & 0.423  & 0.181  & 0.098  \\ \hline
Joint position in real (rad)   & 0.618  & 0.426  & 0.177  & 0.095  \\ \hline
Difference (sim to real) (rad) & 0.001  & -0.003 & 0.004  & 0.003  \\ \hline
Percent error                  & 0.16\% & 0.70\% & 2.26\% & 3.16\% \\ \hline
\end{tabular}
\vspace{-2mm}
\end{table}

Our second experiment is fixing different links on the cable horizontally to the world in simulation and real-world to compare the deformation of the cable-tip. We compare the rotation angles around the $z$-axis from frame we have fixed to the world to frame ``tag\_1''. This experiment is to compare the sagging angle of the cable-tip in simulation and real-world. Similar to the first experiment, three comparisons are made for validation.


Table~\ref{table:validation_no_mass} shows the comparison result by fixing different links on the cable to the world. From the last row of Table~\ref{table:validation_no_mass}, we conclude that the difference of the sagging angle between the simulation model and real cable is within a reasonable range ($<$~0.04 $rad$), and the largest percent error is 3.38$\%$.

\begin{table}[ht]
\centering
\caption{Comparison of the cable deformation between simulation and real world (no external force)}
\label{table:validation_no_mass}
\begin{tabular}{|c|c|c|c|c|}
\hline
Fixed link id / tag id & 5/5   & 4/4   & 3/3   & 2/2   \\ \hline
Sagging angle in sim (rad)         & 1.181 & 1.009 & 0.694 & 0.478 \\ \hline
Sagging angle in real (rad)  & 1.176 & 0.976 & 0.685 & 0.466 \\ \hline
Difference (sim to real) (rad)      & 0.005 & 0.033 & 0.009 & 0.012 \\ \hline
Percent error (\%)  & 0.43\% & 3.38\% & 1.31\% & 2.58\% \\ \hline
\end{tabular}
\vspace{-3mm}
\end{table}

\begin{table}[ht]
\centering
\caption{Comparison of the cable deformation between simulation and real world (adding a 50~$g$ weight to the tip)}
\label{table:validation_50g}
\begin{tabular}{|c|c|c|c|c|}
\hline
Fixed link id / tag id & 5/5    & 4/4   & 3/3   & 2/2   \\ \hline
Sagging angle in sim (rad)         & 1.301  & 1.142 & 0.789 & 0.529 \\ \hline
Sagging angle in real (rad)  & 1.304  & 1.122 & 0.775 & 0.513 \\ \hline
Difference (sim to real) (rad)      & -0.003 & 0.020 & 0.014 & 0.016 \\ \hline
Percent error (\%)  & 0.23\% & 1.78\% & 1.81\% & 3.12\% \\ \hline
\end{tabular}
\vspace{-2mm}
\end{table}

\begin{table}[ht]
\vspace{2mm}
\centering
\caption{Comparison of the cable deformation between simulation and real world (adding a 100~$g$ weight to the tip)}
\label{table:validation_100g}
\begin{tabular}{|c|c|c|c|c|}
\hline
Fixed link id / tag id & 5/5    & 4/4   & 3/3   & 2/2   \\ \hline
Sagging angle in sim (rad)         & 1.328  & 1.216 & 0.838 & 0.587 \\ \hline
Sagging angle in real (rad)  & 1.354  & 1.168 & 0.822 & 0.573 \\ \hline
Difference (sim to real) (rad)      & -0.026 & 0.048 & 0.016 & 0.014 \\ \hline
Percent error (\%)  & 1.92\% & 4.11\% & 1.95\% & 2.44\% \\ \hline
\end{tabular}
\vspace{-4mm}
\end{table}
We also run this experiment with two different weights (50~$g$ and 100~$g$) added to the cable-tip. Table \ref{table:validation_50g} and Table \ref{table:validation_100g} show the results. In the case of a 50~$g$ weight added to the cable-tip, the maximum difference of the cable sagging angle between the simulation and the real world is 0.02~$rad$, and the largest percent error is 3.12$\%$. If a 100~$g$ weight is added to the cable-tip, the maximum difference of the cable sagging angle between the simulation and the real world is 0.048~$rad$, and the largest percent error is 4.11$\%$.

\section{CONCLUSIONS}
In this paper, we proposed a new strategy called Sim2Real2Sim. This strategy includes two procedures: sim-to-real and real-to-sim. The sim-to-real procedure is to organize the methods used in simulation and apply them to the real world in a reasonable way. The real-to-sim procedure is for updating the models in simulation which bridges the gap between simulation and real-world. A novel identification method based on inverse dynamics is implemented to update the parameters of the simulated flexible objects. This paper also shows the steps we take to automate the DRC Plug Task. The success of the system framework running both in simulation and real-world demonstrates the practicability and reliability of the Sim2Real2Sim strategy. Moreover, the comparison of the deformation of the flexible objects in simulation and real-world proves the accuracy of the model in the simulation. Possible future work includes applying our Sim2Real2Sim strategy to more robot tasks and getting more evaluations on this approach, and updating the simulation model by adding more degrees of freedom (e.g., using more apriltags).

\section*{Acknowledgment}
This research is supported by the National Aeronautics and Space Administration under Grant No. NNX16AC48A issued through the Science and Technology Mission Directorate, by the National Science Foundation under Award No. 1544895, 1928654, 1935337, 1944453  and by the Office of the Secretary of Defense under Agreement Number W911NF-17-3-0004.

\bibliographystyle{IEEEtran}
\bibliography{IEEEabrv,IEEEexample}

\end{document}